\documentclass[final]{cvpr}

\usepackage{times}
\usepackage{epsfig}
\usepackage{graphicx}
\usepackage{amsmath}
\usepackage{amssymb}

\usepackage{amsfonts}
\usepackage{algorithm,algpseudocode}
\usepackage{array}
\usepackage{subfig}

\usepackage{color}

\newcommand{\low}{({\color{green}$\downarrow$})}
\newcommand{\high}{({\color{red}$\uparrow$})}

\usepackage[pagebackref=true,breaklinks=true,colorlinks,bookmarks=false]{hyperref}

\def\cvprPaperID{844} 
\def\confYear{CVPR 2021}

\begin{document}

\title{Training Generative Adversarial Networks in One Stage}

\author{Chengchao Shen\textsuperscript{1}, Youtan Yin\textsuperscript{1}, Xinchao Wang\textsuperscript{2,5}, Xubin Li\textsuperscript{3}, Jie Song\textsuperscript{1,4,*}, Mingli Song\textsuperscript{1}\\
\textsuperscript{1}Zhejiang University, 
\textsuperscript{2}National University of Singapore,
\textsuperscript{3}Alibaba Group,\\
\textsuperscript{4}Zhejiang Lab,
\textsuperscript{5}Stevens Institute of Technology\\
{\tt\small \{chengchaoshen,youtanyin,sjie,brooksong\}@zju.edu.cn,}\\
{\tt\small xinchao@nus.edu.sg,lxb204722@alibaba-inc.com}
}

\maketitle

\renewcommand{\thefootnote}{*}
\footnotetext{Corresponding author}
\thispagestyle{empty} 

\begin{abstract}
Generative Adversarial Networks (GANs) have demonstrated unprecedented
success in various image generation tasks.
The encouraging results, however, come at the price of a
cumbersome training process, during which the generator and
discriminator are alternately updated in two stages.
In this paper, we investigate a general training scheme
that enables training GANs efficiently in only one stage.
Based on the adversarial losses of the 
generator and discriminator, we categorize GANs into two classes,
Symmetric GANs and Asymmetric GANs, 
and introduce a novel gradient decomposition method 
to unify the two, allowing us to train both classes
in one stage and hence alleviate the training effort.
We also computationally analyze the efficiency of the proposed method, and empirically demonstrate that,
the proposed method yields a solid $1.5\times$
acceleration across various datasets and 
network architectures.
Furthermore, we show that the proposed method is 
readily applicable to other 
adversarial-training scenarios, 
such as data-free knowledge distillation.
The code is available at \url{https://github.com/zju-vipa/OSGAN}. 
\end{abstract}

\section{Introduction}
Generative Adversarial Networks (GANs),
since their introduction in~\cite{goodfellow2014generative},
have produced unprecedentedly impressive results
on various image generation tasks. Thanks to
the adversarial nature of the two key components,
generator and discriminator, 
the synthesized images delivered by GANs
turn out visually appealing
and in many cases indistinguishable from real ones. 
Recently, many variants of GANs have introduced
and focused on different aspects of the design, including
image quality~\cite{karras2018progressive,brock2019large,donahue2019large},
training stability~\cite{martin2017wasserstein,zhao2017energy,miyato2018spectral}
and diversity~\cite{che2016mode,yang2019diversity,mao2019mode}.
Apart from generating images as the end goal, 
GANs have also been applied to other tasks, such as 
data-free knowledge distillation~\cite{micaelli2019zero,fang2019data,shen2020progressive,Ye_CVPR_2020} 
and domain adaption~\cite{ganin2015unsupervised,saito2018maximum}.

\begin{figure}[t]
    \centering
    \includegraphics[width=0.85\linewidth]{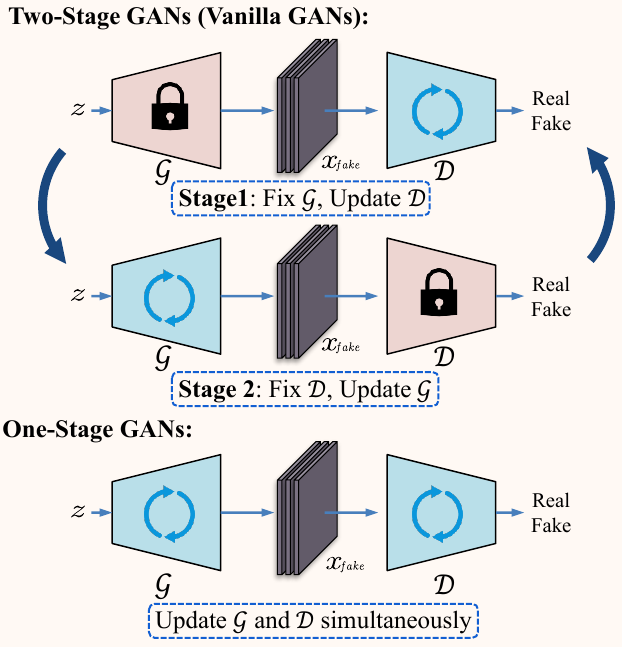}
    \vspace{-1em}
    \caption{Comparison of the conventional Two-Stage GAN training scheme~(TSGANs)
    and the proposed One-Stage strategy~(OSGANs).
    The former one relies on alternately 
    freezing the generator and the discriminant,
    while the latter trains both simultaneously.
    }
    \label{fig:motivation}
    \vspace{-1em}
\end{figure}

The promising results delivered by GANs, however,
come at the price of a burdensome training process. 
As shown in the upper row of Fig.~\ref{fig:motivation}, 
existing GANs rely on a time-consuming two-stage training process,
which we term as \emph{Two-Stage GANs}~(TSGANs).
In the first stage, fake images synthesized by the generator,
together with the real ones,
are fed into the discriminator for training;
during this process, the discriminator is updated but the generator is fixed.
In the second stage, the discriminator delivers the gradients derived from 
the loss function 
to the generator, during which the generator is updated but the discriminator is fixed.
Within each adversarial round, therefore,
both the generator and the discriminator carry out 
the feed-forward step for two times, 
while the discriminator implements a backward-propagation step 
for another two times, 
which, as will be analyzed in our method section,
involves many repetitive computations.  

Endeavors have been made towards alleviating the 
cumbersome training process of GANs.
The work of~\cite{ganin2015unsupervised},
for example, adopts an efficient adversarial training 
strategy for unsupervised domain adaption, 
where learning the feature extractor and classifier 
requires only one round of 
forward inference and back-propagation.
The approach of~\cite{nowozin2016f} also
exploits a single-step optimization 
to update the parameters of the
generator and discriminator in one turn,
and showcase its power in generating visually realistic
images.

In spite of their enhanced efficiency,
the approaches of~\cite{ganin2015unsupervised,nowozin2016f}
limit themselves applicable to only a subset of GANs,
for which their loss functions take a particular form.
Specifically, within such GANs, the adversarial loss terms
in both the generator and discriminator are identical; 
hence, we term such models as \emph{Symmetric GANs}.
Nevertheless, many other popular GANs
adopt loss functions that hold different 
adversarial terms for the generator and discriminator,
and we term these models as \emph{Asymmetric GANs}.
The speed-up optimization techniques employed by~\cite{ganin2015unsupervised,nowozin2016f},
unfortunately, are no longer competent
to handle such asymmetric models.

We propose in this paper a novel one-stage training
scheme, termed as \emph{One-Stage GANs}~(OSGANs),
that generalizes to both Symmetric and Asymmetric GANs.
Our key idea is 
to integrate the optimization for 
generator and discriminator 
during forward inference,
and decompose their gradients during back-propagation
to respectively update them in one stage.
For the Symmetric case, 
since the discriminant loss 
hold a term that is identical to the generator loss, 
we only need to compute 
the gradient of this term once and 
adopt it for both losses.
In this way, the updates of the generator 
and discriminator may safely take place
in one forward and backward step.

Training Asymmetric GANs {is} more tricky
since we can no longer  
copy the gradients derived from 
the {discriminator} to the generator. 
To this end, we carefully look into  
the composition of the discriminator's gradients.
We discover that,  the gradients 
derived from the different adversarial terms,
in reality,
preserve their proportions within the total gradients 
from the last layer all the way back to
the first layer of discriminator. 
This interesting property of gradients,
in turn, provides us with a feasible solution to 
decompose the gradients of the different adversarial terms
and then to update the discriminator and generator,
enabling the one-stage training of Asymmetric GANs. 
Finally, we unify the two classes of GANs, and show that
Symmetric GANs, in fact, can be treated as 
a degenerate case of Asymmetric GANs.

Our contribution is therefore a general 
one-stage training scheme,
readily applicable to various GAN variants
including both Symmetric and Asymmetric GANs.
Computational analysis backed up with experimental results
on several datasets and network architectures
demonstrate that, 
the proposed OSGANs achieve
a {solid} $1.5\times$ speedup over
the vanilla adversarial training strategy.


\section{Related Work}
We briefly review here two lines of work 
related to ours, including GANs
and those one-stage or two-stage
frameworks for other vision tasks.

\subsection{Generative Adversarial Networks}
The pioneering work of GAN~\cite{goodfellow2014generative} 
introduces an adversarial strategy, where generator is trained to synthesize fake images to confuse discriminator and discriminator tries to distinguish synthesized images from real one. 
When the generator and discriminator converge to a competitive equilibrium, the generator can finally synthesize realistic images from latent variables. 

{The mainstream GAN research
has been focused on improving
image quality~\cite{karras2018progressive,brock2019large,donahue2019large},
training stability~\cite{martin2017wasserstein,zhao2017energy,miyato2018spectral}
and diversity~\cite{che2016mode,yang2019diversity,mao2019mode}.}
Recently, GANs have
demonstrated their promising results
in various image-generation tasks, 
such as super resolution~\cite{ledig2017photo,wang2018esrgan,bell2019blind,Soh_2019_CVPR,wang2018sftgan}, image editing~\cite{zhu2016generative,brock2017neural,zhu2020indomain,gu2019mask,he2019attgan}, image inpainting~\cite{pathak2016context,iizuka2017globally,Qiu_ECCV2020},
and image translation~\cite{isola2017image,zhu2017unpaired,tung2017adversarial}.

\subsection{One-Stage and Two-Stage Framework}
Recent object detection methods 
can be categorized into 
one-stage  and two-stage frameworks~\cite{liu2020deep}.
For two-stage framework~\cite{girshick2014rich,girshick2015fast,ren2015faster,dai2016r}, 
a category-agnostic region proposal module is implemented to find the possible object locations in the image, and then a category-specific classifier assigns class label for each location, where multiple feed forwards are implemented. 
One-stage detection methods~\cite{sermanet2014overfeat,liu2016ssd,redmon2016you,law2018cornernet},
on the other hand, 
unify the object bounding box and class prediction in one feed forward, which significantly reduce computation cost. 

Similar to object detection, 
instance segmentation can also be grouped into one-stage and two-stage pipeline. 
For two-stage framework~\cite{pinheiro2015learning,he2017mask,liu2018path,chen2018masklab,chen2019tensormask}, the model first performs object detection to obtain bounding box for each object instance, and then implements binary segmentation inside each bounding box to obtain final results.
With the similar motivation as detection, one-stage instance segmentation~\cite{wang2020solov2,bolya2019yolact} unifies instance mask prediction and category prediction, which effectively improves inference efficiency. 
Other tasks such as classification, however, have been mainly relied on
one-stage schemes~\cite{Yang_CVPR_2020,Yu_CVPR_2017,Yang_NeurIPS_2020,jing2020dynamic,jing2018stroke}.

The above one-stage methods focus on the improvement 
of inference efficiency for their tasks, yet our proposed method pays more attention to the improvement of training efficiency for GANs. 



\section{Method}
In this section, we describe the proposed 
method for one-stage GANs in detail. 
We first describe
the two classes of GANs,
Symmetric GANs and Asymmetric GANs,
and then discuss the one-stage solutions 
for the above cases, respectively. 
Finally, we  analyze the speedup 
factor of the proposed OSGANs.

\subsection{One-Stage GANs}
\subsubsection{Background}
The vanilla GAN proposed by~\cite{goodfellow2014generative} 
introduces a minmax game between discriminator $\mathcal{D}$ 
and generator $\mathcal{G}$ to 
guide the generator to synthesize realistic images.
The objective is expressed as:
\begin{align}\label{eq:gan}
 \mathop{\min}\limits_\mathcal{G} \! \mathop{\max}\limits_\mathcal{D} \mathop{\mathbb{E}} \limits_{x \sim {p_{d}}} \left[\log \mathcal{D}(x) \right] + \! \mathop{\mathbb{E}} \limits_{z \sim {p_z}} \left[\log (1 - \mathcal{D}(\mathcal{G}(z)))\right], 
\end{align}
where $x$ denotes real sample obtained from data distribution $p_{d}$ and $z$ denotes latent variable sampled from Guassian distribution $p_{z}$.
For convenience, we rewrite the above objective as separated losses for $\mathcal{D}$ and $\mathcal{G}$ as follows:
\begin{align}\label{eq:symmetric}
    \mathcal{L}_\mathcal{D} = -\log \mathcal{D}(x) - \log \left( 1 - \mathcal{D}(\mathcal{G}(z)) \right), \\ 
    \mathcal{L}_\mathcal{G} = \log \left( 1 - \mathcal{D}(\mathcal{G}(z)) \right), \nonumber
\end{align}
where $\mathcal{L}_\mathcal{D}$ and $\mathcal{L}_\mathcal{G}$ have the same adversarial term about $\mathcal{G}(z)$: $\log \left( 1 - \mathcal{D}( \mathcal{G}( z ) ) \right)$. 
Hence, we designate such GANs as \textbf{\emph{Symmetric GANs}}. 

To alleviate the gradient vanishing problem of generator, a non-saturating loss function~\cite{arjovsky2017towards} is proposed as follows:
\begin{align}\label{eq:non-saturate}
    \mathcal{L}_\mathcal{D} = -\log \mathcal{D}(x) - \log \left( 1 - \mathcal{D}(\mathcal{G}(z)) \right), \\ 
    \mathcal{L}_\mathcal{G} =-\log \left( \mathcal{D}(\mathcal{G}(z)) \right), \nonumber
\end{align}
where their adversarial terms about $\mathcal{G} \left(z \right)$ are different. 
We term GANs with such losses as \textbf{\emph{Asymmetric GANs}}.
The above adversarial learning scheme is just one instance. 
Many other adversarial formulas can be found in the supplementary material, such as WGAN~\cite{martin2017wasserstein} and LSGAN~\cite{Mao2017}.

In general, the adversarial terms in GANs' objective are the ones about fake sample $\mathcal{G}(z)$, such as $-\log \left( 1 - \mathcal{D}( \mathcal{G}( z ) ) \right)$ vs $\log \left( 1 - \mathcal{D}( \mathcal{G}( z ) ) \right)$ in Eq.~\ref{eq:symmetric} and $-\log \left( 1 - \mathcal{D}( \mathcal{G}( z ) ) \right)$ vs $-\log \left( \mathcal{D}(\mathcal{G}(z)) \right)$ in Eq.~\ref{eq:non-saturate}. 
Therefore, for clarity of analysis, we split general $\mathcal{L}_\mathcal{D}$ into two parts: the term about real sample $\mathcal{L}_\mathcal{D}^{r}(x)$ and the term about fake sample $\mathcal{L}_\mathcal{D}^{f}(\hat{x})$, where $\hat{x}=\mathcal{G}(z)$.
\footnote{There are other forms of GANs, 
whose $\mathcal{L}_\mathcal{D}$ can not be explicitly split in this way. 
Yet still, our approach is applicable.}
More discussions can be found in the supplementary material. 
Finally, the objective of general GANs can be written as follows:
\begin{align}\label{eq:general_objective}
    \mathcal{L}_\mathcal{D}(x, \hat{x}) = \mathcal{L}_\mathcal{D}^{r}(x) + \mathcal{L}_\mathcal{D}^{f}(\hat{x}), \\ 
    \mathcal{L}_\mathcal{G}(\hat{x}) = \mathcal{L}_\mathcal{G}(\mathcal{G}(z)). \nonumber
\end{align}
For brevity, we omit input $x$ and $\hat{x}$, 
such as $\mathcal{L}_\mathcal{D}$, $\mathcal{L}_\mathcal{D}^{r}$, $\mathcal{L}_\mathcal{D}^{f}$ and $\mathcal{L}_\mathcal{G}$.
Based on Eq.~\ref{eq:general_objective}, 
we can formally distinguish Symmetric GANs and Asymmetric GANs:
for Symmetric GANs, we have $\mathcal{L}_\mathcal{G} = -\mathcal{L}_\mathcal{D}^{f}$,
while for Asymmetric GANs, 
we have $\mathcal{L}_\mathcal{G} \neq -\mathcal{L}_\mathcal{D}^{f}$.

\subsubsection{Symmetric OSGANs}
For Symmetric GANs (e.g. Eq.\ref{eq:symmetric}), $\mathcal{L}_\mathcal{G}$ and $\mathcal{L}_\mathcal{D}$ contain the same loss term about fake sample $\hat{x}$: $\mathcal{L}_\mathcal{D}^{f}$.
Their gradients w.r.t. $\hat{x}$ can be denoted as $\nabla_{\hat{x}} \mathcal{L}_\mathcal{D} = \nabla_{\hat{x}} \mathcal{L}_\mathcal{D}^{f}$ and $\nabla_{\hat{x}} \mathcal{L}_\mathcal{G} = -\nabla_{\hat{x}} \mathcal{L}_\mathcal{D}^{f}$, respectively.
We can obtain $\nabla_{\hat{x}} \mathcal{L}_\mathcal{G}$ from $\nabla_{\hat{x}} \mathcal{L}_\mathcal{D}$ by just multiplying $\nabla_{\hat{x}} \mathcal{L}_\mathcal{D}$ with $-1$ and then compute the gradients w.r.t. parameters of $\mathcal{G}$ from $\nabla_{\hat{x}} \mathcal{L}_\mathcal{G}$. 
In a word, we can update the parameters of $\mathcal{G}$ with $\nabla_{\hat{x}} \mathcal{L}_\mathcal{D}$, which is computed during the training of $\mathcal{D}$. 
This method simplifies the training of Symmetric GANs from two stages to one stage.

\subsubsection{Asymmetric OSGANs}
For Asymmetric GANs (e.g. Eq.\ref{eq:non-saturate}), due to $\mathcal{L}_\mathcal{G} \neq -\mathcal{L}_\mathcal{D}^{f}$, the gradients $\nabla_{\hat{x}} \mathcal{L}_\mathcal{G}$ can not be directly obtained from $\nabla_{\hat{x}} \mathcal{L}_\mathcal{D}$ as Symmetric GANs.

An intuitive idea to solve this problem is to integrate $\mathcal{L}_\mathcal{G}$ and $\mathcal{L}_\mathcal{D}$ into one, e.g. $\mathcal{L}_\mathcal{D} \leftarrow \mathcal{L}_\mathcal{D} + \mathcal{L}_\mathcal{G}$, so that we can obtain $\nabla_{\hat{x}} \mathcal{L}_\mathcal{G}$ from $\nabla_{\hat{x}} \mathcal{L}_\mathcal{D}$.
Since the sign of $\nabla_{\hat{x}} \mathcal{L}_\mathcal{D}^{f}$ and $\nabla_{\hat{x}} \mathcal{L}_\mathcal{G}$ is generally reversed, we adopt 
\begin{equation}\label{eq:original_loss}
\mathcal{L} =  \mathcal{L}_\mathcal{D} - \mathcal{L}_\mathcal{G} = \mathcal{L}_\mathcal{D}^{r} + \mathcal{L}_\mathcal{D}^{f} - \mathcal{L}_\mathcal{G},
\end{equation}
instead of $\mathcal{L} =  \mathcal{L}_\mathcal{D} + \mathcal{L}_\mathcal{G}$ to avoid the gradient counteraction between $\mathcal{L}_\mathcal{D}^{f}$ and $\mathcal{L}_\mathcal{G}$.
We denote $\mathcal{L}_{f} = \mathcal{L}_\mathcal{D}^{f} - \mathcal{L}_\mathcal{G}$ to collect the loss terms about fake sample.
By doing so, however,
another issue is introduced:
how to recover $\nabla_{\hat{x}} \mathcal{L}_\mathcal{G}$ from mixed gradients $\nabla_{\hat{x}} \mathcal{L}_{f}$.

To tackle this problem, 
we investigate back-propagation about the discriminator network. 
We find an interesting property of the mainstream neural modules.
The relation between the gradients of loss $\mathcal{L}$ 
w.r.t. the input $\nabla_{in} \mathcal{L}$,
and the gradient of loss $\mathcal{L}$ 
w.r.t. its output $\nabla_{out} \mathcal{L}$
can be presented as the follows:
\begin{equation}\label{eq:relation}
    \nabla_{in} \mathcal{L} = P \cdot \mathcal{F} \left( \nabla_{out} \mathcal{L} \right) \cdot Q,
\end{equation}
where $P$ and $Q$ are matrices that depend on the module or its input; 
$\mathcal{F}(\cdot)$ is a function that satisfies
the equation $\mathcal{F}(y_1 + y_2) = \mathcal{F}(y_1) + \mathcal{F}(y_2)$. 
The neural modules include convolutional module, 
fully-connected module, non-linear activation operation, 
and pooling.
Although both non-linear activation and pooling are non-linear operations, the relation between their gradients of loss function w.r.t. input and output meets Eq.~\ref{eq:relation}. 
More details are discussed in the supplementary material. 

Based on Eq.~\ref{eq:relation}, we can obtain the relation between the gradients of $\mathcal{L}_\mathcal{G}$ and $\mathcal{L}_\mathcal{D}$ for fake sample $\hat{x}_i$ in discriminator as follows: 
\begin{equation}\label{eq:proportion}
    \frac{\nabla_{\hat{x}_i} \mathcal{L}_\mathcal{G}}{\nabla_{\hat{x}_i} \mathcal{L}_\mathcal{D}}
    = \cdots = \frac{\nabla_{\hat{x}_i^l} \mathcal{L}_\mathcal{G}}{\nabla_{\hat{x}_i^l} \mathcal{L}_\mathcal{D}} 
    = \cdots = \frac{\nabla_{\hat{x}_i^L} \mathcal{L}_\mathcal{G}}{\nabla_{\hat{x}_i^L} \mathcal{L}_\mathcal{D}} 
    = \gamma_i,
\end{equation}
where 
$L$  is the total number of layers, 
$\gamma_i$ is an instance-wise scalar for $\hat{x}_i$,\footnote{The 
derivation can be found in the supplementary material.
In fact, $\gamma_i$ is a tensor with the same size as $\hat{x}_i^l$,
in which all elements have the same value. 
For brevity, we regard it as a scalar.}
and $\hat{x}_i^l$ denotes features of $l$-th layer in discriminator.

In other words, $\gamma_i$ remains
a constant for sample $\hat{x}_i^l$ cross different layers, 
despite each sample holds a different $\gamma_i$.
On the one hand, we only need to compute $\gamma_i$ for the last layer, and
computing ${\nabla_{\hat{x}_i^L} \mathcal{L}_\mathcal{G}} / {\nabla_{\hat{x}_i^L} \mathcal{L}_\mathcal{D}}$ to obtain $\gamma_i$ for all layers
is in practice very efficient.
On the other hand, it varies with different samples and therefore needs to be re-computed for each sample. 
Since $\gamma_i$ is the ratio between two scalars $\nabla_{\hat{x}_i^L} \mathcal{L}_\mathcal{G}$ and $\nabla_{\hat{x}_i^L} \mathcal{L}_\mathcal{D}$, 
the cost of this tiny computation can be
omitted during training.

Combining $\nabla_{\hat{x}_i^l} \mathcal{L}_{f} = \nabla_{\hat{x}_i^l} \mathcal{L}_{\mathcal{D}}^{f} - \nabla_{\hat{x}_i^l} \mathcal{L}_\mathcal{G}$ and Eq.~\ref{eq:proportion}, 
we can proportionally decompose $\nabla_{\hat{x}_i^l} \mathcal{L}_\mathcal{G}$ and $\nabla_{\hat{x}_i^l} \mathcal{L}_\mathcal{D}^{f}$ from mixed gradients: $\nabla_{\hat{x}_i^l} \mathcal{L}_{f}$:
\begin{align}\label{eq:asymmetric-gradient}
\nabla_{\hat{x}_i^l} \mathcal{L}_\mathcal{D}^{f} = \frac{1}{1 - \gamma_i} \nabla_{\hat{x}_i^l} \mathcal{L}_{f}, \\
\nabla_{\hat{x}_i^l} \mathcal{L}_\mathcal{G} = \frac{\gamma_i}{1 - \gamma_i} \nabla_{\hat{x}_i^l} \mathcal{L}_{f}. \nonumber 
\end{align}

In other words, we can obtain $\nabla_{\hat{x}_i^l} \mathcal{L}_\mathcal{G}$ and $\nabla_{\hat{x}_i^l} \mathcal{L}_\mathcal{D}^{f}$ by scaling $\nabla_{\hat{x}_i^l} \mathcal{L}_{f}$. 
For the convenience of computation, we apply the above scaling operation to loss function $\mathcal{L}_{f}$, which is equivalent to Eq.~\ref{eq:asymmetric-gradient}. 
Hence, we can obtain the instance loss functions for discriminator and generator:
\begin{align}\label{eq:asymmetric-instance}
\mathcal{L}_\mathcal{D}^{ins} = \mathcal{L}_\mathcal{D}^{r} + \frac{1}{1 - \gamma_i} \left( \mathcal{L}_\mathcal{D}^{f} - \mathcal{L}_\mathcal{G} \right), \\
\mathcal{L}_\mathcal{G}^{ins} = \frac{\gamma_i}{1 - \gamma_i} \left( \mathcal{L}_\mathcal{D}^{f} - \mathcal{L}_\mathcal{G} \right), \nonumber 
\end{align}
where both $\mathcal{L}_\mathcal{D}^{ins}$ and $\mathcal{L}_\mathcal{G}^{ins}$ contain the same loss term $(\mathcal{L}_\mathcal{D}^{f} - \mathcal{L}_\mathcal{G})$.
In this way, Asymmetric GAN is transformed into a symmetric one. 
Hence, a similar one-stage training strategy adopted in Symmetric OSGANs can be applied. 
The difference is that the gradients from $\mathcal{L}_\mathcal{D}^{ins}$ needs to be scaled by $\gamma_i$ instead of $-1$, since $\nabla_{\hat{x}_i^l} \mathcal{L}_\mathcal{G}^{ins} = \gamma_i \cdot \nabla_{\hat{x}_i^l} \mathcal{L}_\mathcal{D}^{ins}$.

A numerical detail needs to considered 
in Eq.~\ref{eq:asymmetric-instance}:
if $(1 - \gamma_i)$ is very close to zero,
a numerical instability issue occurs. 
Due to the competition between $\mathcal{L}_\mathcal{D}^{f}$ and $\mathcal{L}_\mathcal{G}$, the signs of $\nabla_{x_i^l} \mathcal{L}_\mathcal{D}^{f}$ and $\nabla_{x_i^l} \mathcal{L}_\mathcal{G}$ tend to be reversed. 
Hence, $\gamma_i$ is a negative number, which ensures the numerical stability. 

Let us now take a further look into the 
relation between solutions for Symmetric OSGANs and Asymmetric OSGANs. 
Based on Eq.~\ref{eq:proportion}, when $\mathcal{L}_\mathcal{G} = -\mathcal{L}_\mathcal{D}^{f}$ is met, we have $\gamma_i=-1$.
Combining $\gamma_i=-1$ and Eq.~\ref{eq:asymmetric-instance}, we can obtain $\mathcal{L}_\mathcal{D}^{ins} = \mathcal{L}_\mathcal{D}^{r} + \mathcal{L}_\mathcal{D}^{f}$ and $\mathcal{L}_\mathcal{G}^{ins} = -\mathcal{L}_\mathcal{D}^{f}$, which has the same form as Symmetric OSGAN. 
In other words, our method is a general solution for both Symmetric GANs and Asymmetric GANs. 
The overall algorithm for our proposed approach can be summarized as Algorithm~\ref{alg}.

{
    \setlength{\textfloatsep}{2pt}
    \renewcommand{\algorithmicrequire}{\textbf{Input:}} 
    \renewcommand{\algorithmicensure}{\textbf{Output:}}
    \begin{algorithm}[!t]
        \caption{One-Stage GAN Training Framework}
        \label{alg}
        \begin{algorithmic}[1]
            \Require{Training data $\mathcal{X} = \{x_j\}_{j=1}^{N}$.}
            \Ensure{The parameters of generator $\mathcal{G}$}.
            \State Initialize the parameters of $\mathcal{G}$ and $\mathcal{D}$;
            \For{number of training iterations}
                \State Sample $z \sim \mathcal{N} \left( 0,1 \right)$ to generate fake data $\hat{x}_i$ by $\mathcal{G}$;
                \State Sample real data $x_j$ from $\mathcal{X}$;
                \State Feed $\hat{x}_i$ and $x_j$ into $\mathcal{D}$ to compute $\mathcal{L}_{\mathcal{G}}$ and $\mathcal{L}_{\mathcal{D}}$;
                \State Compute $\nabla_{\hat{x}_i^L} \mathcal{L}_{\mathcal{G}}$ and $\nabla_{\hat{x}_i^L} \mathcal{L}_{\mathcal{D}}$ to obtain $\gamma_i$ by Eq.~\ref{eq:proportion};
                \State Compute $\mathcal{L}_\mathcal{D}^{ins}$ by Eq.~\ref{eq:asymmetric-instance};
                \State Back propagate $\nabla_{\hat{x}_i^L} \mathcal{L}_\mathcal{D}^{ins}$ to obtain $\nabla_{\hat{x}_i} \mathcal{L}_\mathcal{D}^{ins}$;
                \State Obtain $\nabla_{\hat{x}_i} \mathcal{L}_\mathcal{G}^{ins} = \gamma_i \cdot \nabla_{\hat{x}_i} \mathcal{L}_\mathcal{D}^{ins}$ by Eq.~\ref{eq:proportion};
                \State Update $\mathcal{G}$ and $\mathcal{D}$ simultaneously using Adam.
            \EndFor
        \end{algorithmic}
    \end{algorithm}
}

\begin{figure*}[t]
    \centering
    \vspace{-2mm}
    \includegraphics[width=\linewidth]{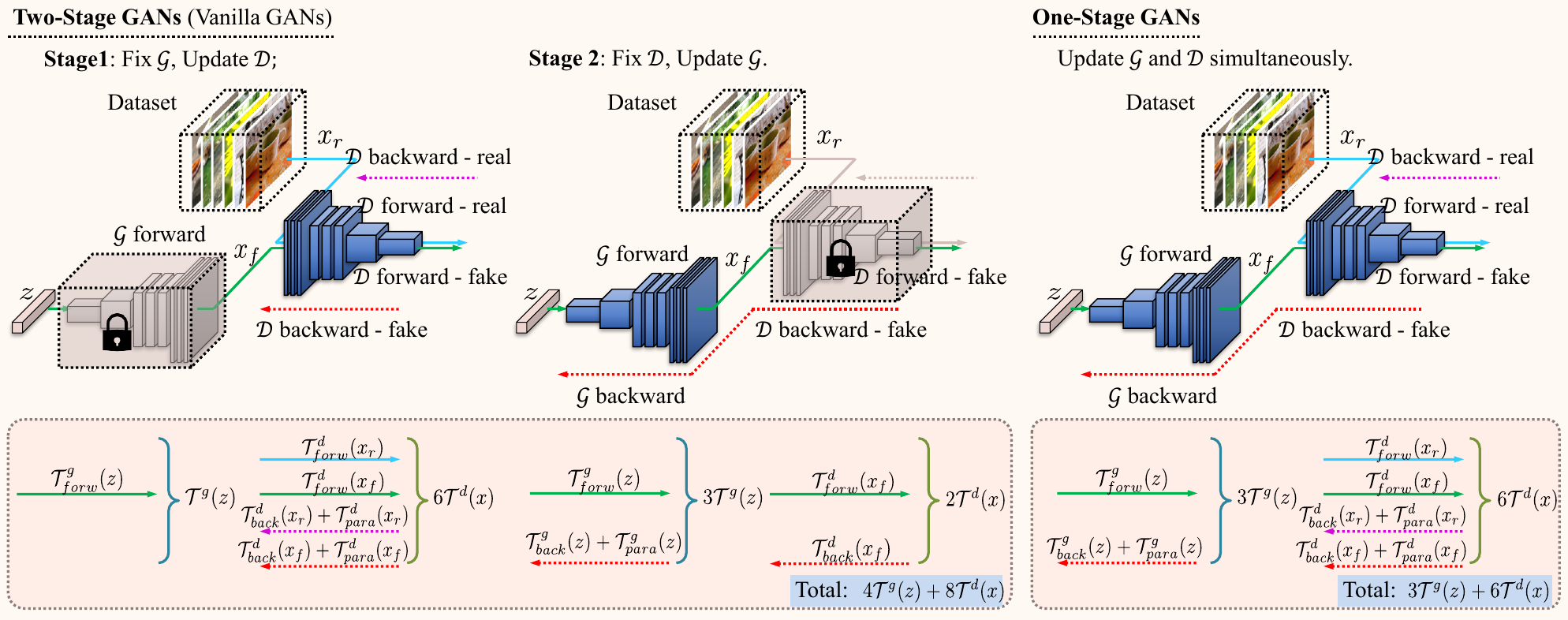}
    \caption{Efficiency comparison between TSGANs (left) and OSGANs (right).
    For TSGANs, in the first stage, $\mathcal{D}$ is trained to classify the fake and real samples, where $\mathcal{G}$ forwards one time and $\mathcal{D}$ forwards and backwards one time for real and fake samples, respectively.
    In the second stage, $\mathcal{G}$ is trained to confuse $\mathcal{D}$, where both $\mathcal{G}$ and $\mathcal{D}$ backward and forward one time. 
    For OSGANs, $\mathcal{G}$ and $\mathcal{D}$ are trained simultaneously, where both of them only need forward and backward one time.
    }
    \vspace{-1em}
    \label{fig:efficiency-analysis}
\end{figure*}

\subsection{Efficiency Analysis}
In this section, we computationally analyze the efficiency of OSGANs and TSGANs. 
To this end, we investigate three perspectives:
(1) the training time for real samples and fake ones in one batch,
(2) the time for forward inference and back-propagation,
{(3) the time for parameter gradient computation and back-propagation.}

For the training time, we assume that the batch size of real samples
and fake ones is equal, which is a widely adopted setting in practice. 
Hence, the time for real samples is equal to the time for fake ones,
which works on both forward inference and back-propagation:
\begin{align}\label{eq:realfake}
    \mathcal{T}_{forw}(x_r) = \mathcal{T}_{forw}(x_f), \\
    \mathcal{T}_{back}(x_r) = \mathcal{T}_{back}(x_f). \nonumber
\end{align}

For the forward and backward time, 
we focus on the cost on modules, 
such as convolutional and fully-connected ones,
which take up the majority of time.
We find the time cost on forward inference 
and back-propagation for these modules 
is approximately equal, which will be 
discussed in the supplementary material. 
For simplicity, the above time is represented as $\mathcal{T}(x)$:
\begin{align}\label{eq:forback}
    \mathcal{T}_{forw}(x) = \mathcal{T}_{back}(x) = \mathcal{T}(x).
\end{align}

For linear modules, such as convolutional
and fully-connected layers,
due to the equal number of floating point operations,
the time consumed on parameter gradient computation and backward is nearly identical.
We write:
\begin{align}\label{eq:upback}
    \mathcal{T}_{para}(x) = \mathcal{T}_{back}(x) = \mathcal{T}(x).
\end{align}

To estimate the low bound of speedup ratio for our method, 
we assume  all  modules in  $\mathcal{G}$ and $\mathcal{D}$  
contain learnable parameters, for which the gradients need computing. 
Under this assumption, the efficiency is analyzed as follow.

As shown in Fig.~\ref{fig:efficiency-analysis}, for TSGANs, 
in the first stage, discriminator is trained to discriminate 
the fake samples synthesized by generator and the real ones.
The generator takes $\mathcal{T}_{forw}^g(z)$ to generate samples.
And the discriminator takes $\mathcal{T}_{forw}^d(x_r) \! + \! \mathcal{T}_{forw}^d(x_f)$ on forward inference,
$\mathcal{T}_{back}^d(x_r) \! + \! \mathcal{T}_{back}^d(x_f)$ on back-propagation
and $\mathcal{T}_{para}^d(x_r) \! + \! \mathcal{T}_{para}^d(x_f)$ on parameter gradients computation.
Combined with Eq.~\ref{eq:realfake}, Eq.~\ref{eq:forback} and Eq.~\ref{eq:upback}, the time for the first stage is
\begin{align}\label{eq:stage1}
\mathcal{T}_{stage1}(z, x) = \mathcal{T}^g(z) + 6\mathcal{T}^d(x),
\end{align} 
where both $\mathcal{T}^g(z)$ and $\mathcal{T}^d(x)$ denote the time cost by generator and discriminator on forward inference and back-propagation, respectively. 

In the second stage, the generator takes
$\mathcal{T}_{forw}^g(z)$, $\mathcal{T}_{back}^g(z)$ and $\mathcal{T}_{para}^g(z)$
on forward inference, back-propagation and parameter gradients computation, respectively. 
The discriminator takes $\mathcal{T}_{forw}^d(x_f)$ and $\mathcal{T}_{back}^d(x_f)$ 
on forward inference and back-propagation, respectively. 
Combined with Eq.~\ref{eq:realfake} and Eq.~\ref{eq:forback}, the time for the second stage is
\begin{align}\label{eq:stage2}
\mathcal{T}_{stage2}(z, x) = 3\mathcal{T}^g(z) + 2\mathcal{T}^d(x).
\end{align} 
Overall, the total time for TSGANs is therefore
\begin{align}\label{eq:two-stage-time}
\mathcal{T}_{two}(z,x) &= \mathcal{T}_{stage1}(z, x) + \mathcal{T}_{stage2}(z, x) \nonumber \\
&=4\mathcal{T}^g(z) + 8\mathcal{T}^d(x).
\end{align}
In the same way, the time for OSGANs is
\begin{align}\label{eq:one-stage-time}
\mathcal{T}_{one}(z,x) = 3\mathcal{T}^g(z) + 6\mathcal{T}^d(x).
\end{align}
Combining Eq.~\ref{eq:two-stage-time} and Eq.~\ref{eq:one-stage-time}, we can obtain the training speedup ratio of OSGANs against TSGANs in the worst case: 
\begin{align}\label{eq:speedup}
    S=\frac{\mathcal{T}_{two}(z,x)}{\mathcal{T}_{one}(z,x)} = \frac{4\mathcal{T}^g(z) + 8\mathcal{T}^d(x)}{3\mathcal{T}^g(z) + 6\mathcal{T}^d(x)} = \frac{4}{3}.
\end{align}

\section{Experiments}
In this section, 
we conduct experiments 
to evaluate the effectiveness of our proposed training method. 
First, we introduce the experimental settings, 
including the datasets we used, evaluation metrics and the implementation details.
We then show
experimental results on popular generation benchmarks 
including quantitative analysis. 
Afterwards, we compare
the training efficiency between one-stage training and two-stage one. 
Finally, we adopt
our one-stage training strategy 
on data-free adversarial distillation.

{
    \begin{table*}[ht]
    \begin{center}
    \resizebox{\linewidth}{!}
    {
        \begin{tabular}{l|cc|cc|cc}
        \hline
        \multicolumn{1}{c|}{} 
            & \multicolumn{2}{c|}{\textbf{CelebA}} 
            & \multicolumn{2}{c|}{\textbf{LSUN Churches}} 
            & \multicolumn{2}{c}{\textbf{FFHQ}}  
            \\ \hline
        \makebox[4cm][c]{\textbf{Method}}  
            & \textbf{FID} & \textbf{KID}  
            & \textbf{FID} & \textbf{KID} 
            & \textbf{FID} & \textbf{KID}
            \\ \hline
        DCGAN~\cite{radford2015unsupervised} (sym, two) 
            & 23.78$\pm$0.12 & 0.019$\pm$0.001 
            & 46.09$\pm$0.25 & 0.043$\pm$0.002 
            & 40.56$\pm$0.32 & 0.040$\pm$0.001 
            \\
        DCGAN (sym, one) 
            & 22.66$\pm$0.09\low & 0.018$\pm$0.001\low 
            & 38.99$\pm$0.21\low & 0.037$\pm$0.001\low 
            & 29.38$\pm$0.24\low & 0.026$\pm$0.001\low 
            \\ 
        \hline
        CGAN~\cite{mirza2014conditional} (sym, two) 
            & 39.85$\pm$0.27 & 0.035$\pm$0.000 
            & / & / & / & / 
            \\
        CGAN (sym, one) 
            & 30.97$\pm$0.33\low & 0.013$\pm$0.000\low 
            & / & / & / & / 
            \\
        \hline
        WGAN~\cite{martin2017wasserstein} (sym, two) 
            & 33.30$\pm$0.23 & 0.028$\pm$0.001 
            & 36.29$\pm$0.14 & 0.034$\pm$0.002 
            & 35.66$\pm$0.13 & 0.032$\pm$0.001 \\
        WGAN (sym, one) 
            & 27.23$\pm$0.18\low & 0.021$\pm$0.001\low 
            & 35.58$\pm$0.30\low & 0.033$\pm$0.002\low 
            & 39.77$\pm$0.14\high & 0.036$\pm$0.001\high \\ 
        \hline
        \hline
        DCGAN~\cite{arjovsky2017towards}$\dag$ (asym, two) 
            & 25.34$\pm$0.20 & 0.022$\pm$0.001 
            & 33.35$\pm$0.40 & 0.030$\pm$0.001 
            & 31.50$\pm$0.15 & 0.031$\pm$0.001 \\
        DCGAN$\dag$ (asym, one) 
            & 24.20$\pm$0.17\low & 0.019$\pm$0.001\low 
            & 25.41$\pm$0.30\low & 0.024$\pm$0.001\low 
            & 31.19$\pm$0.21\low & 0.028$\pm$0.001\low  \\
        \hline
        LSGAN~\cite{mao2017least} (asym, two) 
            & 23.00$\pm$0.17 & 0.019$\pm$0.001 
            & 22.63$\pm$0.16 & 0.021$\pm$0.001 
            & 30.29$\pm$0.23 & 0.029$\pm$0.001 \\
        LSGAN (asym, one) 
            & 19.31$\pm$0.18\low & 0.015$\pm$0.001\low 
            & 22.39$\pm$0.37\low & 0.019$\pm$0.001\low 
            & 29.11$\pm$0.12\low & 0.026$\pm$0.001\low \\ 
        \hline
        GeoGAN~\cite{lim2017geometric} (asym, two) 
            & 21.92$\pm$0.13 & 0.018$\pm$0.001 
            & 18.85$\pm$0.41 & 0.016$\pm$0.001 
            & 30.55$\pm$0.21 & 0.031$\pm$0.001 \\
        GeoGAN (asym, one) 
            & 21.52$\pm$0.14\low & 0.017$\pm$0.001\low 
            & 18.92$\pm$0.24\high & 0.017$\pm$0.001\high 
            & 30.63$\pm$0.26\high & 0.032$\pm$0.001\high \\ 
        \hline
        RelGAN~\cite{jolicoeur2018relativistic} (asym, two) 
            & 20.91$\pm$0.15 & 0.018$\pm$0.001 
            & 24.95$\pm$0.19 & 0.023$\pm$0.002 
            & 35.81$\pm$0.18 & 0.033$\pm$0.001 \\
        RelGAN (asym, one) 
            & 20.79$\pm$0.18\low & 0.015$\pm$0.001\low
            & 25.09$\pm$0.25\high & 0.024$\pm$0.001\high
            & 35.72$\pm$0.16\low & 0.031$\pm$0.001\low  \\
        \hline
        FisherGAN~\cite{mroueh2017fisher} (asym, two) 
            & 27.88$\pm$0.29 & 0.024$\pm$0.001 
            & 29.25$\pm$0.32 & 0.026$\pm$0.001 
            & 52.65$\pm$0.16 & 0.050$\pm$0.001 \\
        FisherGAN (asym, one) 
            & 27.10$\pm$0.19\low & 0.021$\pm$0.001\low 
            & 29.23$\pm$0.18\low & 0.025$\pm$0.001\low 
            & 51.63$\pm$0.18\low & 0.049$\pm$0.001\low  \\
        \hline
        BGAN~\cite{hjelm2018boundary} (asym, two) 
            & 31.12$\pm$0.19 & 0.027$\pm$0.001 
            & 35.06$\pm$0.31 & 0.035$\pm$0.001
            & 41.35$\pm$0.17 & 0.038$\pm$0.001  \\
        BGAN (asym, one) 
            & 25.34$\pm$0.17\low & 0.022$\pm$0.001\low 
            & 34.42$\pm$0.21\low & 0.035$\pm$0.001\low 
            & 42.64$\pm$0.13\high & 0.039$\pm$0.001\high  \\
        \hline
        SNGAN~\cite{miyato2018spectral} (asym, two) 
            & 34.95$\pm$0.24 & 0.033$\pm$0.001 
            & 33.51$\pm$0.16 & 0.034$\pm$0.001 
            & 52.46$\pm$0.24 & 0.050$\pm$0.001  \\
        SNGAN (asym, one) 
            & 34.34$\pm$0.12\low & 0.033$\pm$0.001\low  
            & 31.23$\pm$0.17\low & 0.031$\pm$0.001\low 
            & 51.03$\pm$0.36\low & 0.048$\pm$0.001\low  \\
        \hline
        \end{tabular}
    }

    \caption{Comparative results of OSGANs and TSGANs.
    All results are averaged over five runs 
    and error bars correspond to the standard deviation. 
    Specifically, ``sym'' and ``asym'' respectively 
    denote Symmetric GANs and Asymmetric GANs,
    while ``one'' and ``two'' 
    respectively
    denote one-stage and two-stage strategy.
    $\dag$ adopts asymmetric adversarial loss in~\cite{arjovsky2017towards};
    {\color{green}{$\downarrow$}} 
    and {\color{red}{$\uparrow$}}
    respectively denote 
    our strategy outperforms and underperforms
    the two-stage one.
    }
    \label{table:perfomance}
    \vspace{-2em}
    \end{center}
    \end{table*}
}

\subsection{Experimental Settings}
{
    \subsubsection{Datasets}
    CelebFaces Attributes Dataset (CelebA)~\cite{liu2015faceattributes} is a large-scale face attributes dataset, which consists of more than 200K celebrity images. 
    In the experiment, the images are roughly aligned according to the two eye locations and then resized and cropped into $64\times64$ size.

    LSUN churches~\cite{yu15lsun} is a dataset that contains more than 126K church outdoor images.
    Flickr-Faces-HQ (FFHQ)~\cite{karras2019style} is a face dataset, which consists of 70K images and covers large variation in term of age, ethnicity, image background and accessory.
    For training convenience, all images of the above dataset are resized and cropped into $64\times64$ resolution. 

    Both CIFAR10 and CIFAR100~\cite{krizhevsky2009learning} are composed of 60,000 colour images with 32$\times$32 size, where 50,000 images are used as training set and the rest 10,000 images are used as test set. 
    The CIFAR10 dataset contains 10 classes, and the CIFAR100 contains 100 classes. 
    Random cropping and random horizontal flipping are applied to the training images to augment the dataset. 

    \subsubsection{Implementation Details}
    The proposed method is implemented using PyTorch v1.5 on a Quadro P5000 16G GPU.
    The batch size for both real images and fake one is set to 128.
    In all experiments, Adam optimization algorithm is adopted for both $\mathcal{G}$ and $\mathcal{D}$. 
    For the convenience of experiments, we adopt the network architecture used by DCGANs as the backbone. 
     Given that Eq.~\ref{eq:relation} does not work on batch normalization~\cite{ioffe2015batch}, we replace the batch normalization in $\mathcal{D}$ with group normalization~\cite{wu2018group} so as to meet Eq.~\ref{eq:relation}.

    \subsubsection{Evaluation Metrics}
    To evaluate the performance of GANs, three popular evaluation metrics are available. 

    Fr\'echet Inception Distance (FID)~\cite{heusel2017gans} is used to evaluate the distance between features from real images and the ones from generated images, which is computed as follows:
    \begin{align}
    \operatorname{FID} = \left\|\mu_{r}-\mu_{g}\right\| \! + \! \operatorname{tr}\left(C_{r}+C_{g}-2\left(C_{r} C_{g}\right)^{1 / 2}\right), 
    \end{align}
    where $\mu_{r}$ and $\mu_{g}$ are the empirical means for real and generated samples, $C_{r}$ and $C_{g}$ are the empirical covariance for real and generated samples, respectively. 

    Kernel Inception Distance (KID)~\cite{binkowski2018demystifying} mitigates the overfitting problem, which may occur in FID.
    It is computed by
    \begin{align}\label{eq:kid}
    \operatorname{KID} = \! \mathop{\mathbb{E}}\limits_{\substack{x_r, x_g}} \! 
    \left[k\left(x_{r}, x_{r}^{\prime}\right) \! - \! 2 k\left(x_{r}, x_{g}\right) \! + \! k\left(x_{g}, x_{g}^{\prime}\right)\right], 
    \end{align}
    where $x_r$ and $x_g$ are features for real and generated samples, respectively.
    $k\left(x, y\right) = \left(\frac{1}{d}x^{\operatorname{T}} y  + 1\right)^3$.
    
\begin{figure*}[t]
    \resizebox{\linewidth}{!}
    {
    \begin{tabular}{m{0.2cm}<{\centering}  m{8cm}<{\centering} m{8cm}<{\centering} }

    & \textbf{\small CelebA} & \textbf{ \small LSUN Churches}  \\

    \rotatebox{90}{{\scriptsize LSGAN (two)}} 
    & \includegraphics[width=\linewidth]{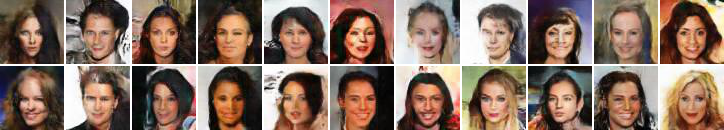}
    & \includegraphics[width=\linewidth]{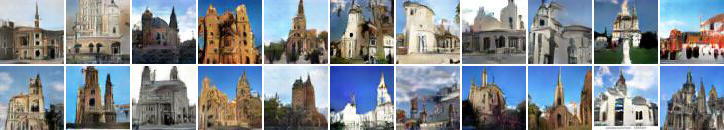} \\

    \rotatebox{90}{{\scriptsize LSGAN (one)}} 
    & \includegraphics[width=\linewidth]{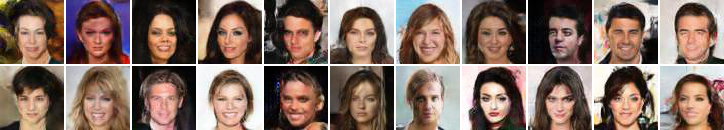}
    & \includegraphics[width=\linewidth]{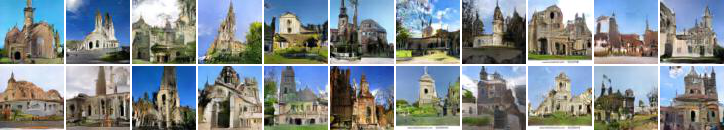} \\

    \hline 

    \rotatebox{90}{{\scriptsize RelGAN (two)}} 
    & \vspace{1mm} \includegraphics[width=\linewidth]{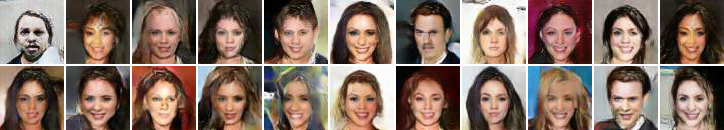}
    & \vspace{1mm} \includegraphics[width=\linewidth]{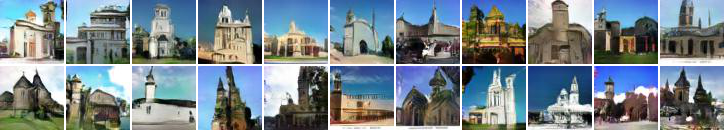} \\

    \rotatebox{90}{{\scriptsize RelGAN (one)}} 
    &\includegraphics[width=\linewidth]{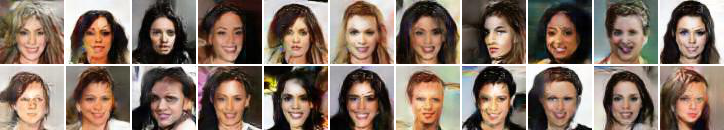}
    &\includegraphics[width=\linewidth]{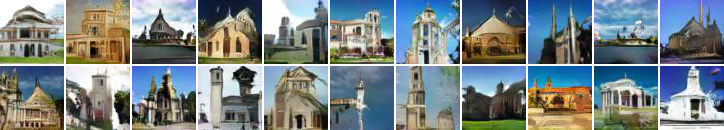} \\

    \end{tabular}
    }

    \caption{
    Results synthesized by TSGANs (``two'') and OSGANs (``one'') on CelebA and LSUN Churches. 
    }
    \vspace{-1em}
    \label{fig:visual}

\end{figure*}

    Inception Score (IS)~\cite{salimans2016improved} is proposed to measure the realistic level
    of a generated image.
    As discussed in~\cite{barratt2018note} and~\cite{rosca2017variational}, applying the IS to generative models trained on datasets other than ImageNet gives misleading results. 
    The IS should only be used as a ``rough guide'' to evaluate generative models.
    Hence, in the following comparative experiments, we will focus on FID and KID.
}

\subsection{One-Stage Adversarial Training}
{
    To verify the effectiveness of one-stage adversarial training, 
    we compare the performance on several representative GANs, 
    each of which is trained with one-stage strategy and two-stage one, respectively. 
    The comparative results,
    including those obtained by 
    symmetric GANs and asymmetric GANs,
    are shown in Tab.~\ref{table:perfomance},

    Symmetric GANs trained with one-stage training strategy,
    in reality, outperform their two-stage counterparts, 
    except for WGAN on FFHQ. 
    We analyze a possible reason as follows.
    In two-stage training, the gradients of 
    generator are computed after the training of discriminator.
    This asynchronous gradient computation may lead to
    inefficient optimization, where the earlier optimization 
    information is lost during the training of generator. 
    The one-stage training, on the other hand,
    effectively bypasses the problem by synchronously
    updating the generator and discriminator. 

    Variants of Asymmetric GANs with different 
    asymmetric adversarial objectives are evaluated. 
    On DCGAN, LSGAN, FisherGAN and SNGAN, our one-stage strategy 
    consistently outperforms their two-stage counterparts on all datasets. 
    The performance of one-stage GeoGAN, RelGAN and BGAN
    is on par with their corresponding two-stage one. 
    This can be in part explained by that, 
    due to the adversarial-objective difference between generator and discriminator,
    the effect of optimization information loss is not so evident. 
    Overall, our one-stage strategy training achieves  performance 
    comparable to those of the existing two-stage ones. 
    Some visualization results are shown in Fig.~\ref{fig:visual}. 
    More results can be found in the supplementary material.
}

\begin{figure*}[t]
    \vspace{-3mm}
    \renewcommand\arraystretch{0}
    \renewcommand\tabcolsep{0pt}
    \resizebox{\linewidth}{!}
    {
    \begin{tabular}{m{0.5cm}<{\centering} m{8cm}<{\centering} m{8cm}<{\centering} m{8cm}<{\centering}}

     & \textbf{CelebA} & \textbf{LSUN Churches} & \textbf{FFHQ}  \\

    \rotatebox{90}{DCGAN(sym)} 
    & \includegraphics[width=\linewidth]{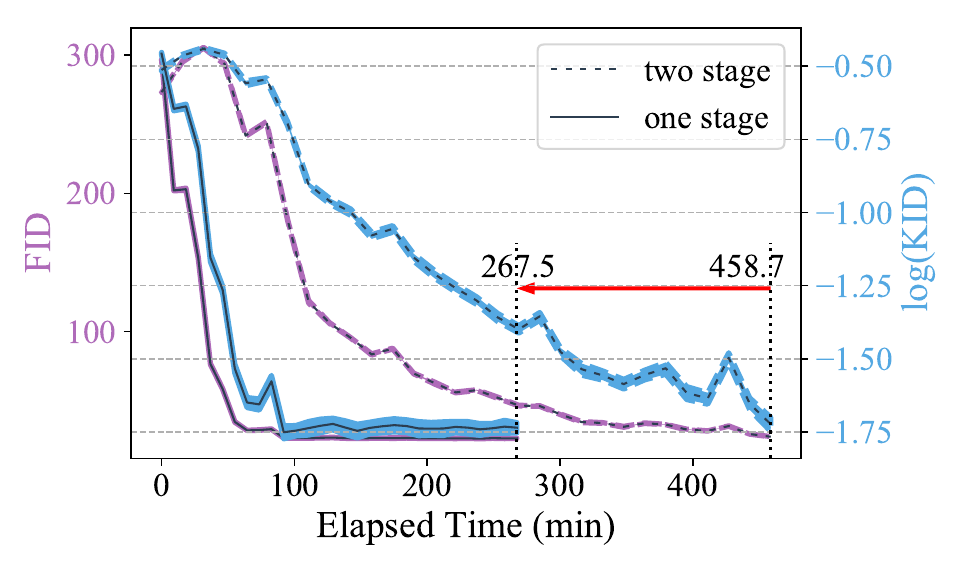}
    & \includegraphics[width=\linewidth]{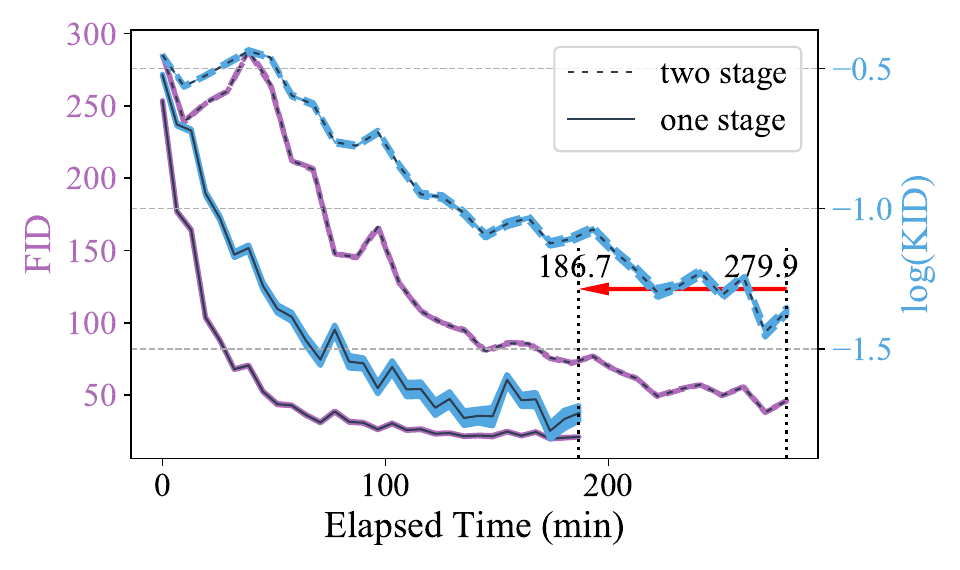}
    & \includegraphics[width=\linewidth]{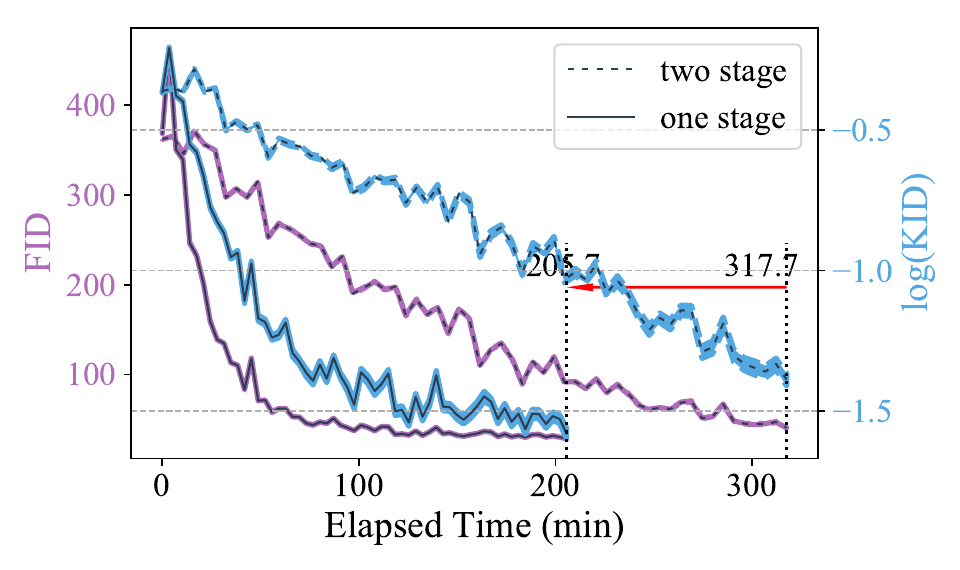} \\


    \rotatebox{90}{DCGAN(asym)} 
    & \includegraphics[width=1\linewidth]{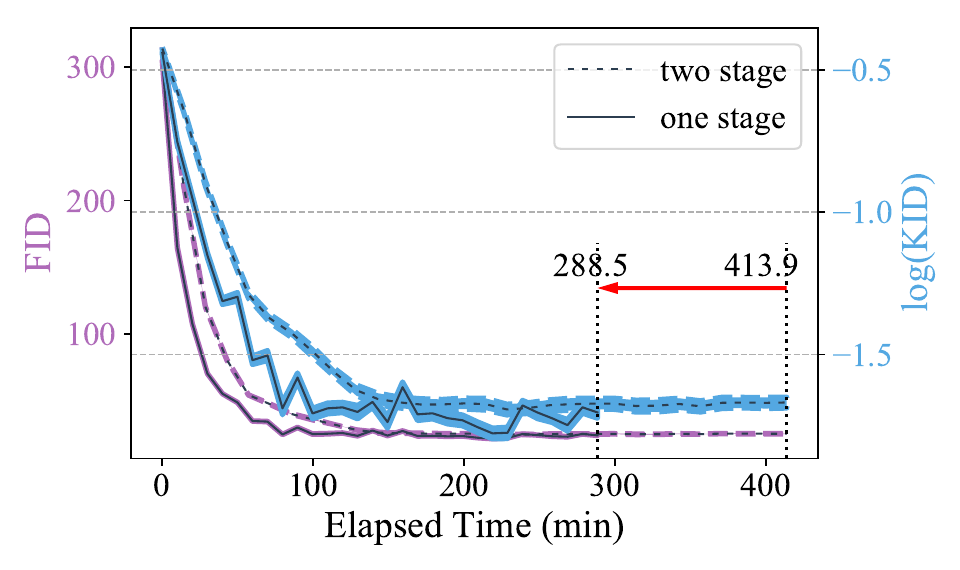}
    & \includegraphics[width=1\linewidth]{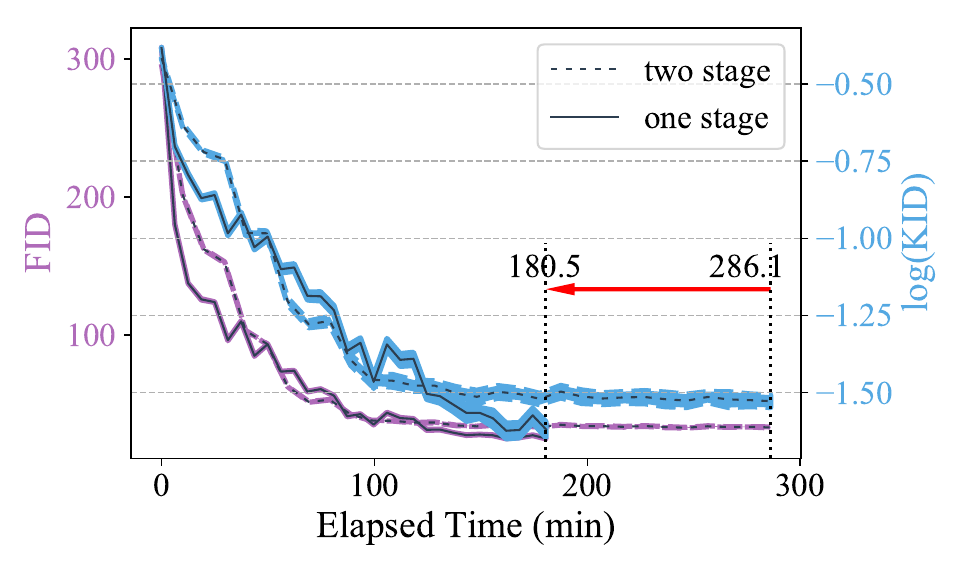}
    & \includegraphics[width=1\linewidth]{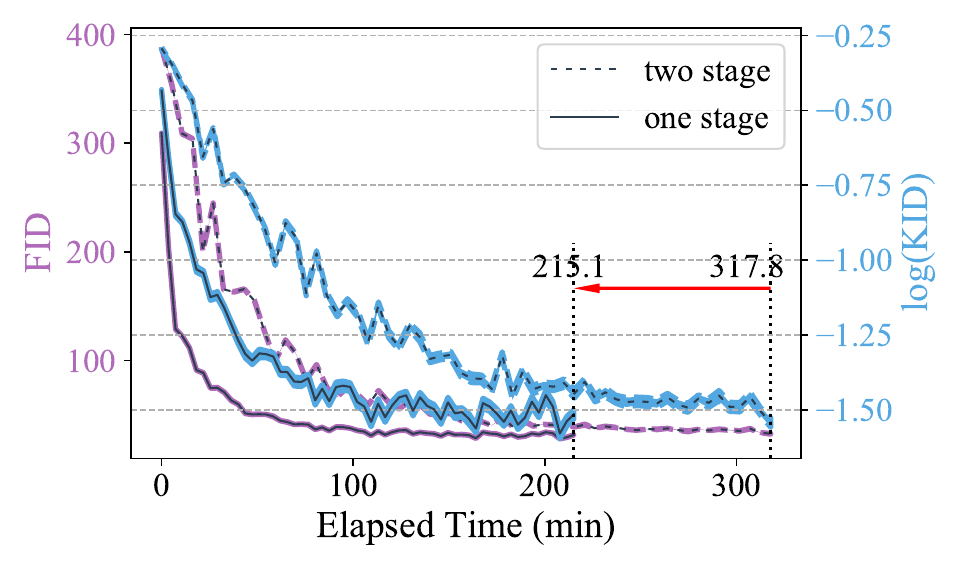} \\


    \rotatebox{90}{LSGAN(asym)} 
    & \includegraphics[width=\linewidth]{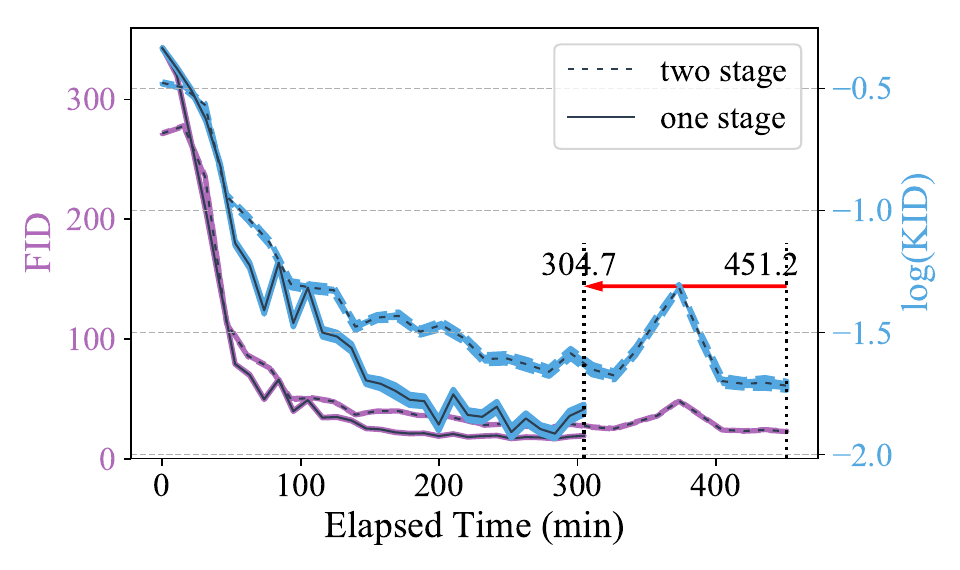}
    & \includegraphics[width=\linewidth]{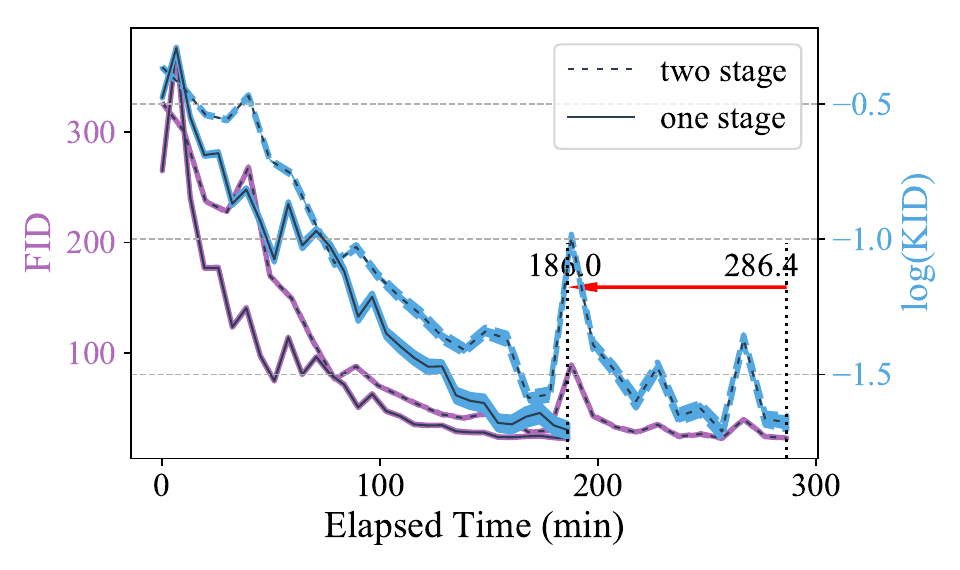}
    & \includegraphics[width=\linewidth]{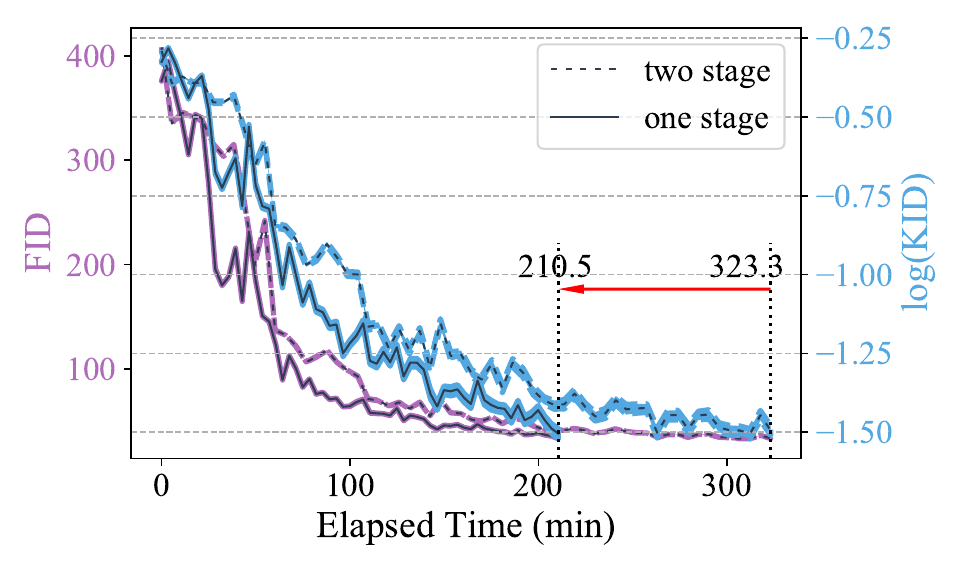} \\

    \end{tabular}
    }

    \caption{
    Comparison of training efficiency between OSGANs and TSGANs. 
    All experiments are terminated when their performance reaches plateau.
    }
    \label{fig:efficiency}
    \vspace{-1.5em}
\end{figure*}

\subsection{Efficiency Analysis}
{
    In this section, we empirically evaluate the efficiency of OSGANs against TSGANs. 
    As shown in Fig.~\ref{fig:efficiency}, both Symmetric GAN (DCGAN-sym) and Asymmetric GAN (DCGAN-asym and LSGAN-asym) are compared with the corresponding TSGANs, respectively. 
    Due to small value of KID, we take logarithm of KID for better visualization.

    For Symmetric DCGAN, one-stage training strategy
    tends to be more stable than two-stage one. 
    For Symmetric GAN, the end-to-end training strategy significantly smooths the gradients for optimization, which leads to a more stable training process. 
    One-stage Symmetric DCGAN outperforms the two-stage one on all datasets, while it achieves a speedup ratio sometimes greater than 1.5
    (e.g. $458.7/267.5 \approx 1.7$). 
    We believe that one-stage training adopts the synchronous update for the generator and discriminator, which can significantly 
    avoid the early gradient saturation in two-stage strategy.

    For asymmetric DCGAN, one-stage strategy achieves faster convergence speed and better performance than two-stage one on CelebA and LSUN Churches 
    (speedup ratio: $286.1/180.5 \approx 1.6$). 
    For asymmetric LSGAN, one-stage training consistently converges faster than the two-stage one on all datasets (speedup ratio: $286.4/186.0 \approx 1.5$).
    In the meantime, one-stage training has smaller fluctuation than the two-stage one.
    More results can be found in the supplementary material.
}

\subsection{Application: Knowledge Distillation}
{
    We also investigate the effectiveness of our training 
    strategy on adversarial knowledge distillation task~\cite{shen2019amalgamating,shen2019customizing}.
    Both DFAD~\cite{fang2019data} and ZSKT~\cite{micaelli2019zero} adopt an adversarial training strategy to implement knowledge distillation without real data, where teacher is ResNet34 and the student is ResNet18.
    Specifically, in each adversarial round, the student  
    learns to imitate the prediction of the teacher by five iterations, 
    while the generator tends to synthesize harder samples
    with larger loss values by one iteration. 
    In other words, it adopts a two-stage adversarial training strategy. 
    Using our proposed method, 
    the adversarial distillation achieves competition within one stage. 
    There is only one update on generator and student network in each round, respectively. 
    Experimental results are shown in Tab.~\ref{table:distill}, demonstrating that our proposed method 
    achieves  performance on par with the original two-stage one, 
    sometimes even better.
    Especially on CIFAR100, our method outperforms  DFAD and ZSKT by a large margin,  significantly reducing the performance gap between knowledge distillation using full dataset and data-free knowledge distillation. 
}

{
    \begin{table}[ht]
    \begin{center}
    \vspace{-3mm}
    \resizebox{\linewidth}{!}
    {
        \begin{tabular}{c|cc|cc}
        \hline
        \multicolumn{1}{c|}{} & \multicolumn{2}{c|}{\textbf{CIFAR10}} & \multicolumn{2}{c}{\textbf{CIFAR100}} \\ \hline
        \textbf{Method}  & \textbf{\#Params} & \textbf{Accuracy} & \textbf{\#Params}  & \textbf{Accuracy} \\ \hline 
        Teacher 
            & 21.3M & 0.955 
            & 21.3M & 0.775 \\ \hline
        KD~\cite{hinton2015distilling} 
            & 11.2M & 0.939 
            & 11.2M & 0.733 \\ \hline
        ZSKT~\cite{micaelli2019zero} (two) 
            & 11.2M & 0.921 
            & 11.2M & 0.662 \\ 
        DFAD~\cite{fang2019data} (two) 
            & 11.2M & 0.933 
            & 11.2M & 0.677 \\ 
        Ours(one) 
            & 11.2M & \textbf{0.934} 
            & 11.2M & \textbf{0.700} \\ \hline
        \end{tabular}
    }
    \caption{Comparative results of 
    one-stage adversarial knowledge distillation 
    and two-stage one.
    Here, ``one'' and ``two'' denote one-stage and two-stage strategy, respectively.}
    \label{table:distill}
    \vspace{-8mm}
    \end{center}
    \end{table}

}

\section{Conclusion}
{
    In this paper, we investigate a general one-stage training strategy
    to enhance the training efficiency of GANs. 
    We categorize GANs into two classes,
    Symmetric GANs and Asymmetric GANs. 
    For Symmetric GANs, 
    the gradient 
    computation of generator is obtained from the back-propagation of discriminator during the training. 
    The parameters of both generator and discriminator 
    can therefore be updated in one stage. 
    For Asymmetric GANs, we propose a gradient decomposition method, 
    which integrates the asymmetric adversarial losses during
    forward inference,
    and decomposes their gradients during 
    back-propagation to separately update generator and discriminator in one stage. 
    We analyze the relation between the above 
    solutions and unify them into 
    our one-stage training strategy.
    Finally, we computationally analyze the training 
    speedup ratio for OSGANs against TSGANs. 
    {Experimental results demonstrate that 
    OSGANs achieve more than $1.5\times$ speedup
    over TSGANs,
    and meanwhile preserve the results of TSGANs. }
}

\paragraph{Acknowledgements.}
This work is supported by National Natural Science Foundation of China (U20B2066, 61976186), Key Research and Development Program of Zhejiang Province~(2018C01004),  Major Scientifc Research Project of Zhejiang Lab~(No. 2019KD0AC01) and Alibaba-Zhejiang University Joint Research Institute of Frontier Technologies.

{\small
\bibliographystyle{ieee_fullname}
\bibliography{mybib}
}

\end{document}